\documentclass[11pt,a4paper]{article}
\usepackage[hyperref]{naaclhlt2019}
\usepackage{times}
\usepackage{latexsym}
\usepackage{booktabs}
\usepackage{url}
\usepackage{comment}
\usepackage{multicol}

\newcommand{\example}[1]{\textit{#1}}

\aclfinalcopy 

\title{Training on Synthetic Noise\\Improves Robustness to Natural Noise in Machine Translation}

\author{Vladimir Karpukhin \qquad Omer Levy \qquad Jacob Eisenstein \qquad Marjan Ghazvininejad \\
\\
Facebook Artificial Intelligence Research \\
Seattle, WA
}

\date{}

\begin{document}
\maketitle
\begin{abstract}
We consider the problem of making machine translation more robust to character-level variation at the source side, such as typos. Existing methods achieve greater coverage by applying subword models such as byte-pair encoding (BPE) and character-level encoders, but these methods are highly sensitive to spelling mistakes. We show how training on a mild amount of random synthetic noise can dramatically improve robustness to these variations, without diminishing performance on clean text. We focus on translation performance on natural noise, as captured by frequent corrections in Wikipedia edit logs, and show that robustness to such noise can be achieved using a balanced diet of simple synthetic noises at training time, without access to the natural noise data or distribution.

\end{abstract}

\section{Introduction}

Machine translation systems are generally trained on clean data, without spelling errors. Yet machine translation may be used in settings in which robustness to such errors is critical: for example, social media text in which there is little emphasis on standard spelling~\cite{michel2018mtnt}, and interactive settings in which users must enter text on a mobile device. Systems that are trained on clean data generally perform poorly when faced with such errors at test time~\cite{heigold2017robust,belinkov2018synthetic}.

One potential solution is to introduce noise at training time, an approach that is similar in spirit to the use of adversarial examples in other areas of machine learning~\cite{goodfellow2014explaining} and natural language processing~\cite{ebrahimi-EtAl:2018:Short}. 
So far, using synthetic noise at training time has been found to only improve performance on test data with exactly the same kind of synthetic noise, while at the same time impairing performance on \emph{clean} test data~\cite{heigold2017robust,belinkov2018synthetic}.
We desire methods that yield good performance on both clean text as well as naturally-occurring noise, but this is beyond the reach of current techniques.

Drawing inspiration from dropout \cite{srivastava2014dropout} and noise-based regularization methods,
we explore the space of random noising methods at training time, and evaluate performance on both clean text and text corrupted by natural noise based on real spelling mistakes on Wikipedia~\cite{max2010mining}. We find that by feeding our translation models a balanced diet of several types of synthetic noise at training time -- random character deletions, insertions, substitutions, and swaps -- it is possible to obtain substantial improvements on such naturally noisy data, with minimal impact on the performance on clean data, and without accessing the test noise data or even its distribution.

We demonstrate that our method substantially improves the robustness of a transformer-based machine translation model with CNN character encoders to spelling errors across multiple input languages (German, French, and Czech).
Of the different noise types we use at training, we find that random character deletions are particularly useful, followed by character insertions.
However, noisy training does not seem to improve translations of social media text, as indicated by performance on the recently-introduced MTNT dataset of Reddit posts~\cite{michel2018mtnt}. This finding aligns with previous work arguing that the distinctive feature of social media text is not noise or orthographical errors, but rather, variation in writing style and vocabulary~\cite{eisenstein2013bad}.

\begin{table*}[t]
\centering
\begin{tabular}{lll}
\toprule
\textbf{Deletion} & A character is deleted. & whale $\to$ whle \\
\textbf{Insertion} & A character is inserted into a random position. & whale $\to$ wxhale \\
\textbf{Substitution} & A character is replaced with a random character. & whale $\to$ whalz \\
\textbf{Swap} & Two adjacent characters change position. & whale $\to$ wahle \\
\bottomrule
\end{tabular}
\caption{The synthetic noise types applied during training. Noise is applied on a random character, selected from a uniform distribution. The right column illustrates the application of each noise type on the word ``whale''.}
\label{tab:synthetic_noise}
\end{table*}

\section{Noise Models}

We focus on \emph{orthographical noise}, which is noise at the character level, affecting the spelling of individual terms.
Orthographical noise is obviously problematic for machine translation systems that operate on token-level embeddings, because noised terms are likely to be out-of-vocabulary, even when pre-segmented into subwords using techniques such as byte-pair encoding~\cite{sennrich2015neural}. 
A more subtle issue is that orthographical noise can also pose problems for \emph{character-level} encoding models. 
Typical character-level encoders are based on models such as convolutional neural networks~\cite{kim2016character}, which learn to match filters against specific character n-grams. When these n-grams are disrupted by orthographical noise, the resulting encoding may be radically different from the encoding of a ``clean'' version of the same text. 
\newcite{belinkov2018synthetic} report significant degradations in performance after applying noise to only a small fraction of input tokens.\footnote{Decreased performance is not necessarily an artifact of CNNs; \newcite{belinkov2018synthetic} show similar results on RNN-based character encoders as well, and we have also observed these results in preliminary experiments with bag-of-character models (not reported).}

\subsection{Synthetic Noise}
\label{sec:synthetic_noise}

Table~\ref{tab:synthetic_noise} describes the four types of synthetic orthographic noise we used during training. Substitutions and swaps were experimented with extensively in previous work \cite{heigold2017robust,belinkov2018synthetic}, but deletion and insertion were not. Deletion and insertion pose a different challenge to character encoders, since they alter the distances between character sequences in the word, as well as its overall length. In Section~\ref{sec:ablation}, we show that they are indeed the primary contributors in improving our model's robustness to natural noise.

During training, we used a balanced diet of all four noise types by sampling the noise, for each token, from a multinomial distribution of 60\% clean (no noise) and 10\% probability for each type of noise. The noise was added dynamically, allowing for different mutations of the same example over different epochs.\footnote{We did not apply noise to single-character tokens. Swaps were not applied to the first and last characters of a token.}

\subsection{Natural Noise}
\label{sec:natural_noise}

In addition to these forms of synthetic noise, we evaluate our models on natural noise, which is obtained by mining edit histories of Wikipedia in search of local error corrections (for French and German)~\cite{max2010mining,zesch2012measuring} and manually-corrected essays (for Czech) \cite{sebesta2017}. 
Specifically, these authors have obtained a set of likely spelling error pairs, each involving a clean spelling and a candidate error.
These errors are then injected into the source language.
When there are multiple error forms for a single term, an error is selected randomly (uniformly).
Not all words have error candidates, and so even when noise is applied at the maximum rate, only 20-50\% of all tokens are noised, depending on the language.

Natural noise is more representative of what might actually be encountered by a deployed machine translation system, so we reserve it for test data.
While it is also possible, in theory, to use natural noise for training as well, it is not always realistic. For one, significant engineering effort is required to obtain such noise examples, making it difficult to build naturally-noised training sets for any source language. Secondly, the kind of spelling mistakes may vary greatly across different demographics and periods, making it difficult to anticipate the exact distribution of noise one might encounter at test time.

\section{Experiment}

\begin{table*}[t]
\centering
\begin{tabular}{cccccc}
\toprule
& \multicolumn{2}{l}{} & \multicolumn{3}{c}{\textbf{BLEU}}\\
\cmidrule(lr){4-6}
\textbf{Dataset} & \textbf{Noise Probability} & \textbf{Noised Tokens} & \textbf{Clean Training Data} & \textbf{+ Synthetic Noise} & $\Delta$ \\
\midrule
de-en & ~~~~0.00\% & ~~0.00\% & 34.20 & 33.53 & --0.67\\
de-en & ~~~~6.25\% & ~~2.44\% & 32.73 & 32.95 & ~~0.22\\
de-en & ~~25.00\% & ~~9.72\% & 27.93 & 31.32 & ~~3.39\\
de-en & 100.00\% & 39.36\% & 12.49 & 23.34 & 10.85\\[6pt]
fr-en & ~~~~0.00\% & ~~0.00\% & 39.61 & 39.94 & ~~0.33\\
fr-en & ~~~~6.25\% & ~~3.32\% & 37.34 & 38.49 & ~~1.15\\
fr-en & ~~25.00\% & 13.47\% & 30.48 & 34.07 & ~~3.59\\
fr-en & 100.00\% & 53.74\% & 11.48 & 19.43 & ~~7.95\\[6pt]
cs-en & ~~~~0.00\% & ~~0.00\% & 27.48 & 27.09 & --0.39\\
cs-en & ~~~~6.25\% & ~~1.56\% & 26.63 & 26.51 & --0.12\\
cs-en & ~~25.00\% & ~~6.14\% & 24.31 & 24.91 & ~~0.60\\
cs-en & 100.00\% & 24.53\% & 16.64 & 18.91 & ~~2.27\\
\bottomrule
\end{tabular}
\caption{The performance of a machine translation model on the IWSLT 2016 task when different amounts of natural orthographical errors are introduced. Noise Probability is the probability of a test-data token to be injected with natural noise, while the Noised Tokens column reflects the number of tokens that were actually distorted in practice; not every word in the vocabulary has a corresponding misspelling.
Training on synthetic noise significantly increases robustness in scenarios where large amounts of spelling mistakes are encountered.}
\label{tab:main_result}
\end{table*}

\paragraph{Data}
Following \newcite{belinkov2018synthetic}, we evaluated our method on the IWSLT 2016 machine translation benchmark \cite{junczys2016university}. We translated from three source languages (German, French, Czech) to English. Synthetic noise was only added to the training data, and natural noise was only added to the test data; the validation data remained untouched. 

\paragraph{Model}
We used a transformer-based machine translation model \cite{vaswani2017attention} with the CNN-based character encoder of \newcite{kim2016character} on the source encoder. The model was implemented in Fairseq.\footnote{\url{https://github.com/pytorch/fairseq}}

\paragraph{Hyperparameters}
We followed the base configuration of the transformer \cite{vaswani2017attention}, with 6 encoder and decoder layers of 512 model dimension, 2048 hidden dimensions, and 8 attention heads per layer. Our character embeddings had 256 dimensions and the character CNN's filters followed the specifications of \newcite{kim2016character}. 
We optimized the model with Adam and used the inverse square-root learning rate schedule typically used for transformers, but with a peek learning rate of 0.001. Each batch contained a maximum of 8,000 tokens. We used a dropout rate of 0.2. 
We used beam search for generating the translations (5 beams), and computed BLEU scores to measure performance on the test set.

\subsection{Results}

Table~\ref{tab:main_result} shows the performance of the model on data with varying amounts of natural orthographical errors (see Section~\ref{sec:natural_noise}). As observed in prior art \cite{heigold2017robust,belinkov2018synthetic}, when there are significant amounts of natural noise, the model's performance drops significantly. However, training on our synthetic noise cocktail greatly improves performance, regaining between 20\% (Czech) and 50\% (German) of the BLEU score that was lost to natural noise. Moreover, the negative effects of training on synthetic noise seem to be limited to both negative and positive fluctuations that are smaller than 1 BLEU point.

\begin{table}[t]
\centering
\begin{tabular}{lcc}
\toprule
\textbf{Training Noise} & \textbf{BLEU} & $\Delta$ \\
\midrule
No Training Noise & 12.49 & \\
$+$ Deletion & 17.39 & ~~4.90 \\
$+$ Insertion & 15.00 & ~~2.51 \\
$+$ Substitution & 11.99 & --0.50 \\
$+$ Swap & 14.04 & ~~1.55 \\[6pt]
All Training Noise & 23.34 & \\
$-$ Deletion & 14.96 & --8.38 \\
$-$ Insertion & 18.81 & -4.53 \\
$-$ Substitution & 20.23 & --3.11 \\
$-$ Swap & 23.07 & --0.27 \\
\bottomrule
\end{tabular}
\caption{Ablation analysis for IWSLT 2016 de-en, with maximal natural noise in the test set. The top half displays experiments with only one type of noise, while the bottom half shows results with three out of four noise types.}
\label{tab:ablation}
\end{table}

\subsection{Ablation Analysis}\label{sec:ablation}

To determine the individual contribution of each type of synthetic noise, we conduct an ablation study. We first add only one type of synthetic noise at 10\% (i.e. 90\% of the training data is clean), and measure performance. We then take the full set of noise types, and remove a single type at each time to see how important it is given the other noises. 

Table~\ref{tab:ablation} shows the model's performance on the German-to-English dataset when training with various mixtures of noise. We find that deletion is by far the most effective synthetic noise in preparing our model for natural orthographical errors, followed by insertion. The French and Czech datasets exhibit the same trend.

We conjecture that the importance of deletion and insertion is that they distort the typical distances between characters, requiring the CNN character encoder to become more invariant to unexpected character movements. The fact that we use deletion and insertion also explains why our model was able to regain a significant portion of its original performance when confronted with natural noise at test time, while previous work that trained only on substitutions and swaps was not able to do so \cite{belinkov2018synthetic}.

\begin{table}[t]
\centering
\begin{tabular}{ccc}
\toprule
\textbf{Dataset} & \textbf{Clean Train} & \textbf{+ Synthetic Noise} \\
\midrule
en-fr & 21.1 & 20.6 \\
fr-en & 23.6 & 24.1 \\
\bottomrule
\end{tabular}
\caption{The performance of a machine translation model on the MTNT task. Training on synthetic noise does not appear to have a significant impact.}
\label{tab:mtnt}
\end{table}

\section{Translating Social Media Text}

We also apply our synthetic noise training procedure to translation of social media, using the recently-released MTNT dataset of Reddit posts~\cite{michel2018mtnt}, focusing on the English-French translation pair. 
Note that no noise was inserted into the test data in this case; the only source of noise is the non-standard spellings inherent to the dataset.
As shown in \autoref{tab:mtnt}, noised training has minimal impact on performance. 
We did not exhaustively explore the space of possible noising strategies, and so these negative results should be taken only as a preliminary finding. 
Nonetheless, there are reasons to believe that synthetic training noise may not help in this case.
\newcite{michel2018mtnt} note that the rate of spelling errors, as reported by a spell check system, is not especially high in MTNT; other differences from standard corpora include the use of entirely new words and names, terms from other languages (especially English), grammar differences, and paralinguistic phenomena such as emoticons. 
These findings align with prior work showing that social media does not feature high rates of misspellings~\cite{rello2012social}. Furthermore, many of the spelling variants in MTNT have very high edit distance (e.g., \example{catholique} $\to$ \example{catho} [Fr]). It is unlikely that training with mild synthetic noise would yield robustness to these variants, which reflect well-understood stylistic patterns rather than random variation at the character level.

\section{Related work}

The use of noise to improve robustness in machine learning has a long history~\cite[e.g.,][]{holmstrom1992using,wager2013dropout}, with early work by \newcite{bishop1995training} demonstrating a connection between additive noise and regularization. To achieve robustness to orthographical errors, we require noise that operates on the sequence of characters. \newcite{heigold2017robust} demonstrated that synthetic noising operations such as random swaps and replacements can significantly degrade performance when inserted at test time; they also show that some robustness can be obtained by inserting the same synthetic noise at training time. Similarly, the impact of speech-like noise is explored by \newcite{sperber2017toward}. 

Most relevant for us is the work of \newcite{belinkov2018synthetic}, who evaluated on natural noise obtained from Wikipedia edit histories~\cite[e.g.,][]{max2010mining}. They find that robustness to natural noise can be obtained by training on the same noise model, but that (a) training on synthetic noise does \emph{not} yield robustness to natural noise at test time, and (b) training on natural noise significantly impairs performance on clean text. In contrast, we show that training on the right kind and the right amount of synthetic noise can yield substantial improvements on natural noise at test time, without significantly impairing performance on clean data. Our ablation results suggest that deletion and insertion noise --- which were not included by Belinkov and Bisk --- are essential to achieving robustness to natural noise.  

An alternative approach is to build robustness to character permutations directly into the design of the character-level encoder. \newcite{belinkov2018synthetic} experiment with a bag of characters, while \newcite{sakaguchi2017robsut} use character RNNs combined with special representations for the first and last characters of each token. These models are particularly suited for specific types of swapping and scrambling noises, but are not robust to natural noise. We also conducted preliminary experiments with similar noise-invariant models, but found that training a CNN with synthetic noise to work better.

\section{Conclusion}

In this work we take a step towards addressing the challenge 
of making machine translation robust to character-level noise. 
We show how training on synthetic character-level noise, similar in spirit to dropout, can significantly improve a translation model's robustness to natural spelling mistakes. In particular, we find that deleting and inserting random characters play a key role in preparing the model for test-time orthographic variations. While our method works well on typos, it does not appear to generalize to non-standard text in social media. We conjecture that spelling mistakes constitute a relatively small part of the deviations from standard text, and that the main challenges in this domain stem from other linguistic phenomena.

\bibliography{references}
\bibliographystyle{acl_natbib}

\end{document}